\documentclass[runningheads]{llncs}

\usepackage{makeidx}  
\usepackage{graphicx}
\usepackage{subfigure}
\usepackage{multirow}
\usepackage{color,soul} 
\usepackage{amsmath}
\usepackage{amssymb,amsfonts}
\usepackage{physics} 
\usepackage{cases}
\newtheorem{assumption}{Assumption}

\renewcommand{\div}{\nabla\cdot\,}
\newcommand{\bfv}{{\bf v}}

\newcommand{\bfrho}{{\boldsymbol \rho}}
\newcommand{\bfS}{{\bf S}}
\newcommand{\bfI}{{\bf I}}
\newcommand{\bfA}{{\bf A}}

\makeatletter
 \def\SOUL@hlpreamble{%
 \setul{}{2.4ex}
 \let\SOUL@stcolor\SOUL@hlcolor
 \SOUL@stpreamble
 }
\makeatother

\begin{document}
\mainmatter              

\title{GlymphVIS: Visualizing Glymphatic Transport Pathways Using Regularized Optimal Transport}
\titlerunning{GlymphVIS}  
%

\author{Rena Elkin\inst{1}{\let\thefootnote\relax\footnote{{\hspace{-4mm} Corresponding author email: rena.elkin@stonybrook.edu}}} \and Saad Nadeem\inst{2} \and Eldad Haber\inst{3} \and Klara Steklova\inst{3} \and Hedok Lee\inst{4} \and Helene Benveniste\inst{4} \and Allen Tannenbaum\inst{1,5}}

\authorrunning{Elkin \emph{et al}.} 

\tocauthor{Rena Elkin, Saad Nadeem, Eldad Haber, Klara Steklova, Hedok Lee, Helene Benveniste and Allen Tannenbaum}

\institute{Department of Applied Mathematics and Statistics, Stony Brook University\\
\and
Department of Medical Physics, Memorial Sloan Kettering Cancer Center\\
\and
Department of Mathematics, University British Columbia, Vancouver\\
\and
Department of Anesthesiology, Yale School of Medicine\\
\and
Department of Computer Science, Stony Brook University
}

\maketitle              

\begin{abstract}
The glymphatic system (GS) is a transit passage that facilitates brain metabolic waste removal and its dysfunction has been associated with neurodegenerative diseases such as Alzheimer's disease. The GS has been studied by acquiring temporal contrast enhanced magnetic resonance imaging (MRI) sequences of a rodent brain, and tracking the cerebrospinal fluid injected contrast agent as it flows through the GS. We present here a novel visualization framework, GlymphVIS, which uses regularized optimal transport (OT) to study the flow behavior between time points at which the images are taken. Using this regularized OT approach, we can incorporate diffusion, handle noise, and accurately capture and visualize the time varying dynamics in GS transport. Moreover, we are able to reduce the registration mean-squared and infinity-norm error across time points by up to a factor of 5 as compared to the current state-of-the-art method. Our visualization pipeline yields flow patterns that align well with experts' current findings of the glymphatic system.
\end{abstract}
\section{Introduction}
The glymphatic system (GS) is the structural entity whereby waste products are transported from the brain and into lymphatic vessels located outside, in the meninges and along the neck vasculature \cite{nedergaard2013garbage}. Importantly, the GS also flushes out of the brain soluble amyloid beta (A$\beta$) and tau proteins, the main culprits of Alzheimer's disease (AD) in humans and animals \cite{iliff2012paravascular}. Despite the potential implications of the GS for AD and other neurodegenerative conditions, there are significant gaps in our understanding of the waste clearance mechanisms and the physical forces controlling transport.

Glymphatic transport behavior can be observed with a temporal series of contrast enhanced MR images of the rodent brain. Briefly, the small molecular weight gadolinium (Gd) contrast agent (tracer) is infused into the cerebrospinal fluid (CSF) reservoir of the cisterna magna and its spatial distribution into the brain is captured by the successive acquisition of 3D T1-weighted images (234$\mu$m resolution) every $\sim\!\!4$ minutes, for a total of $\sim\!\!3$ hours \cite{lee2018quantitative}. However, these MRIs do not provide directional information on the tracer movement between time points. Therefore, there is an urgent need for a mathematical framework that can capture and help visualize the dynamic tracer behavior in a manner aligning well with the biological understanding.

In this work, we present a novel visualization framework, GlymphVIS, for studying glymphatic transport pathways using regularized optimal transport (OT). The theory of OT seeks the most feasible way to redistribute mass from one given distribution to another while  minimizing the associated cost of transportation (\cite{rr},\cite{villani1}). OT has been used for registration and connectivity analysis of brain white matter \cite{omtEx1}, image morphing \cite{haker}, and has recently been extended to the case of measures of different total mass \cite{peyre}.

\begin{figure}[!t]
\centering
  \includegraphics[width=\textwidth]{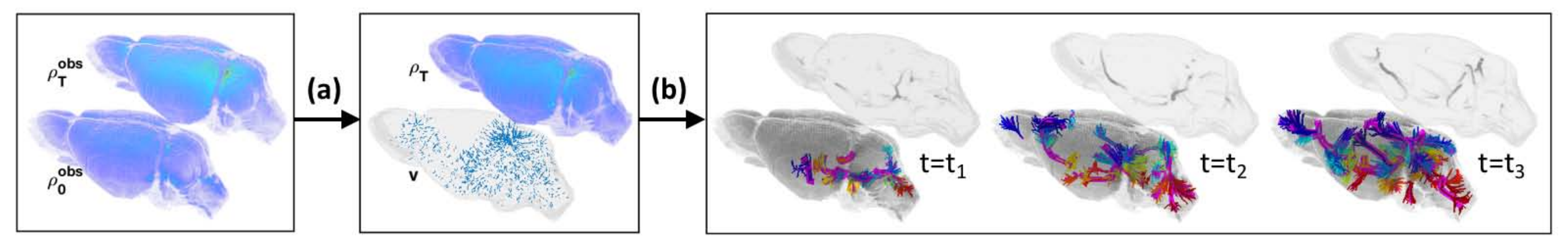}
  \caption{GlymphVIS Pipeline: \textbf{(a)} The generalized regularized OT procedure (GR-OT) takes initial and final `observed' density images as input and returns the `clean' or `believed true' final density image along with the corresponding velocity vector field describing the deformation. \textbf{(b)} The output density images and velocity are then subsequently passed to the flow pattern analysis procedure (FPA) which returns pathway and streamline clustering visualization for the whole time domain.   }
  \label{pip}
\end{figure}

Ratner \emph{et al}. \cite{v1} modeled the glymphatic flow using the traditional OT formulation. This approach yielded promising results, but at the expense of some unrealistic assumptions: (1) the movement of the contrast agent is not affected by diffusion, (2) the total mass of the tracer remains constant over time, and (3) the given MRIs represent the true density distribution at that time. From the implementation perspective, the mass conservation constraint requires normalizing the density distribution which can be drastically altered by the presence of noise in the data and additional noise interference is caused by taking the given images as fixed endpoints.
Finally, the authors of \cite{v1} do not explicitly model time which means the resulting deformation field cannot reflect time-varying dynamics. In this work, we introduce a new and more physiologically relevant model inspired by the work of Benamou and Brenier \cite{BB} that relaxes the above-mentioned unrealistic constraints. Specifically, the contributions of this paper are enumerated as follows:
\begin{enumerate}
    \item We replace the continuity equation with the advection-diffusion equation to more accurately model the flow behavior and smoothen the deformation field;
    \item We no longer enforce total mass conservation, so normalization of the density distributions is not needed;
    \item We treat the final time condition as a free endpoint which prevents overfitting to noise;
    \item We explicitly model the time domain, which allows for a direct temporal analysis of the dynamic flow behavior.
\end{enumerate}

This project was supported by AFOSR grant FA9550-17-1-0435), ARO grant (W911NF-17-1-049), grants from National Institutes of Health (1U24CA18092401A1, R01-AG048769), MSK Cancer Center Support Grant/Core Grant (P30 CA008748), and a grant from Breast Cancer Research Foundation (grant BCRF-17-193).

\section{GlymphVIS}
Benamou and Brenier \cite{BB} recast the OT problem in the context of fluid mechanics that explicitly yields a time-interpolant between the two densities. This naturally motivates an ideal framework for studying the glymphatic pathways because it allows for more direct control and variation in modeling its dynamic flow behavior. Here, we introduce the following two terms to the original Benamou and Brenier OT formulation: (1) a regularization term to alleviate the effect of noise and (2) a diffusion term in the standard continuity equation to better model both advection and diffusion in the glymphatic system. We then clusters the streamlines from the resulting velocity field in order to elucidate and visualize the conduits of glymphatic flow and efflux; see Fig.~\ref{pip}.

\subsection{Regularized OT}
In order to motivate our model formulation, we begin with our assumptions about the data:

\begin{assumption}
Image intensity is proportional to tracer mass, \textup{(and we therefore refer to the intensity as mass)}.
\end{assumption}
\begin{assumption}
Tracer is transported via glymphatic pathway, \textup{as supported by experimental findings \cite{iliff2012paravascular}.}
\end{assumption}
\begin{assumption}
Apparent motion of glymphatic transport is governed by the advection-diffusion equation (ADE),
\begin{eqnarray}
\label{addiff}
 \frac{\partial\rho}{\partial t} + \div (\rho v) = \div \sigma^2 \grad \rho,
\end{eqnarray}
\textup{where} $0\leq\rho:[0,T]\times\mathcal{D}\rightarrow\mathbb{R}$ \textup{is a density with compact support,} $\mathcal{D}\subset\mathbb{R}^d,$ $v:[0,T]\times\mathcal{D}\rightarrow\mathbb{R}^d$ \textup{is velocity and} $0 \le \sigma\in\mathbb{R}$ \textup{is diffusivity}.
\end{assumption}
\begin{assumption}
The MR images we are given are noisy observations of the tracer's conditions at time $t=T_i$,
\begin{equation}
\label{ifcond}
\rho(T_i,x) + \epsilon = \rho_{T_1}^{\rm obs}(x), \quad i = 0,...,N,
\end{equation}
\textup{where} $\epsilon$ \textup{is a random Gaussian iid with covariance} $\Sigma$.
\end{assumption}

Given initial and final observations of tracer density $\rho_0^{\rm obs}$ and $\rho_T^{\rm obs}$ at times $t=0$ and $t=T$ respectively,
our goal is to find the velocity field $v$ and the `believed true' or `clean' image $\rho_T$
such that the constraint \eqref{addiff} is satisfied. To this end, we propose to minimize the objective function
\begin{equation}
\begin{aligned}
\label{gu-OT-orig}
{\cal J}[\rho,v] &= \int_0^T\int_{\mathcal{D}}\, \frac{1}{2}\rho \|v\|^2\, dxdt
+\alpha\|\rho(T,x) - \rho_T^{\rm obs}(x)\|_{\Sigma}^2,
\end{aligned}
\end{equation}
subject to the ADE constraint \eqref{addiff} with the initial condition $\rho(0,x) = \rho_0^{\rm obs}(x)$.
Here, we have added the second term to the Benamou and Brenier energy functional \cite{BB} so that noise, which is inherent in all sensor-derived data, is explicitly taken into account. The parameter $\alpha$ weighs the balance between fitting the data and minimizing the energy associated with transporting the mass. We refer to \eqref{gu-OT-orig} as the \textbf{\emph{generalized regularized OT}} problem (GR-OT) and note that supplemental regularization can easily be implemented by adding intermediate densities to help guide the optimization procedure toward more accurate results.

\medskip \noindent
{\bf Remark:}\\
In the original Benamou-Brenier formulation of OT, $\sigma=0$ and endpoint distribution is specified in equation~(\ref{addiff}), and $\alpha=0$ in equation~(\ref{gu-OT-orig}).

\subsubsection{Proposed minimization method}
While it is possible to solve the optimization problem \eqref{gu-OT-orig} as a constrained one, it is straightforward to eliminate the ADE \eqref{addiff} and solve the problem for $v$ alone.
Accordingly, consider solving the PDE for a given $v$, obtaining the smooth, differentiable map
\begin{eqnarray}
\label{eq:mapF}
F(v) = \rho(t,x), \quad t\in[0,1].
\end{eqnarray}

\subsubsection{Discretization}
Suppose the given images are $(n_1\times\ldots\times n_d)$ in size, let $s=n_1 * \ldots * n_d$ denote the total number of voxels, and let $m$ denote the number of time steps such that $m*\delta t = T$. We will use bold font to denote linearized variables.

We use mimetic methods, designed to keep their properties when considering inner products, for the discretization of the problem.
First, we use operator splitting to discretize the ADE \eqref{addiff} as an advection step and diffusion step, independently.
In the first step, we consider the advection equation and solve the problem
$$\frac{\partial\rho}{\partial t} + \div (\rho v) = 0 \quad \rho(t_n,x) = \rho_n.$$
Using a particle in cell (PIC) method, we obtain the discrete equivalent $\bfrho_{n+1}^* = \bfS (\bfv_n) \bfrho_n$
where $\bfS$ is a linear interpolation matrix.
The method is conservative which means that no mass is lost during this step.
For the second step, we consider the diffusion equation and solve the problem
$$ \frac{\partial\rho}{\partial t} = \div \sigma^2 \grad \rho \quad \rho(t_n,x) = \rho_{n+1}^*. $$
Using the backward Euler method, we obtain the discrete equivalent
\begin{eqnarray}
\label{diffusion}
(\bfI - \delta t \bfA) \bfrho_{n+1} =
 \bfrho_{n+1}^*
\end{eqnarray}
where $\bfI$ is the identity matrix and $\bfA$ is a discretization of the diffusion operator $\div \sigma^2 \grad$ on a cell centered grid.
Combining these two steps, we obtain the corresponding  discrete forward problem $(\bfI - \delta t \bfA) \bfrho_{n+1} =
\bfS (\bfv_n) \bfrho_n, n=0,\ldots,m.$
Clearly, the density at any time step depends only on the initial density $\bfrho_0$ and the velocity $\bfv$, allowing us to define the discrete map $F(\bfv)$ \eqref{eq:mapF} that maps the velocity to the density at all times.
Next, defining $\bfrho = [\bfrho_1^{\top},\ldots,\bfrho_{m+1}^{\top}]^{\top}$ and $\bfv = [\bfv_0^{\top},\ldots,\bfv_m^{\top}]^{\top}$, a straightforward discretization of the energy yields
\begin{eqnarray}
\label{intD}
\int_0^T \int_{\mathcal{D}} \rho \|v\|^2 dx\, dt \approx h^d \delta t\bfrho^\top (\bfI_m \otimes \bfA_v) (\bfv \odot \bfv),
\end{eqnarray}
where $h^d$ is the volume of each cell, $\bfI_k$ is the $k \times k$ identity matrix, $\bfA_v$ is a $1 \times d$ block matrix of $\bfI_s$, $\otimes$ denotes the Kronecker product and $\odot$ denotes the Hadamard product. We then solve the discrete optimization problem which now reads
\begin{eqnarray}
\label{optD}
\min && \phi( \bfv) = \frac{1}{2}h^d\,
\delta t \bfrho^\top (\bfI_m \otimes \bfA_v) (\bfv \odot \bfv) +
\alpha  \|\bfrho_n - \bfrho_n^{\rm obs}\|^2 \\
\nonumber
{\rm subject \ to} && \begin{cases}
(\bfI - \delta t \bfA) \bfrho_{n+1} - \bfS (\bfv_n) \bfrho_n = 0  \\
\bfrho_0 = \bfrho_0^{\rm obs}.
\end{cases}
\end{eqnarray}

Note that the objective function is quadratic with respect to $\bfv$ and the interpolation matrix $\bfS$ is linear with respect to $\bfv$ as it contains the weights on the linear interpolation. Following \cite{Klara}, one can use a Gauss-Newton like method to solve the problem.

\subsection{Flow Pattern Analysis}
The time-interpolant of density images and corresponding time-varying velocity vector field $\bfv(t,x)$ directly output by the GR-OT procedure is then fed into our flow pattern analysis procedure (FPA). For each time step, we 
construct streamlines by integrating the velocity field $\bfv$. By looking at the streamline density through each voxel, we get a global visualization of the GS `pathways.' 
In order to supplement this with local information, we cluster the streamlines using the QuickBundles algorithm \cite{quickbundles}. Significant clusters 
provide more information regarding different flow trajectories within different pathways and fluid reservoirs.
Both the pathways and clusters are converted to NIfTI files where they are analyzed by overlaying anatomical masks using Amira software specifically designed for visualization of data in 3D and 4D. We discuss these results in the following section.

\section{Results}
\begin{figure}[t!]
\begin{center}
\setlength{\tabcolsep}{10pt}
\begin{tabular}{cc}
\includegraphics[width=0.4\textwidth]{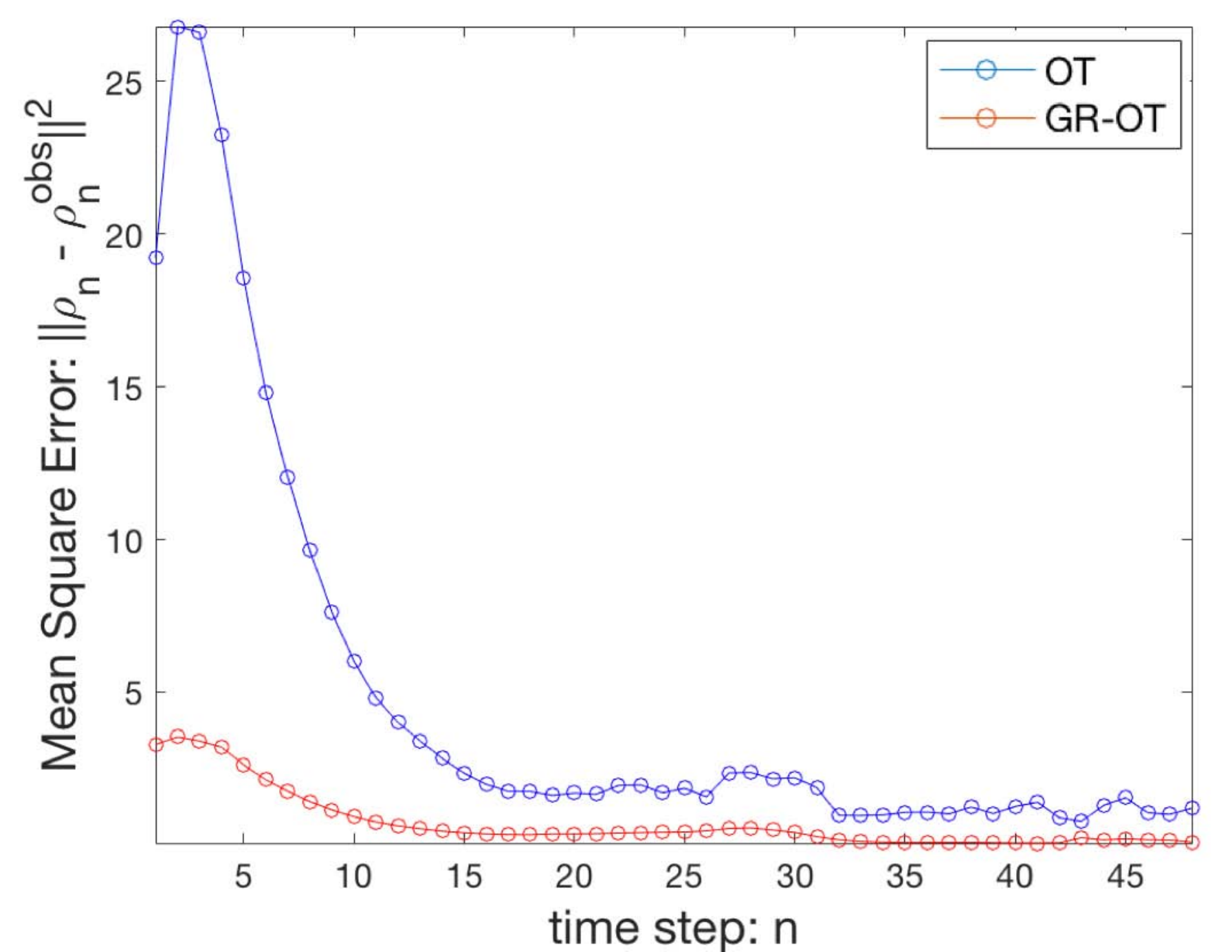}&
\includegraphics[width=0.4\textwidth]{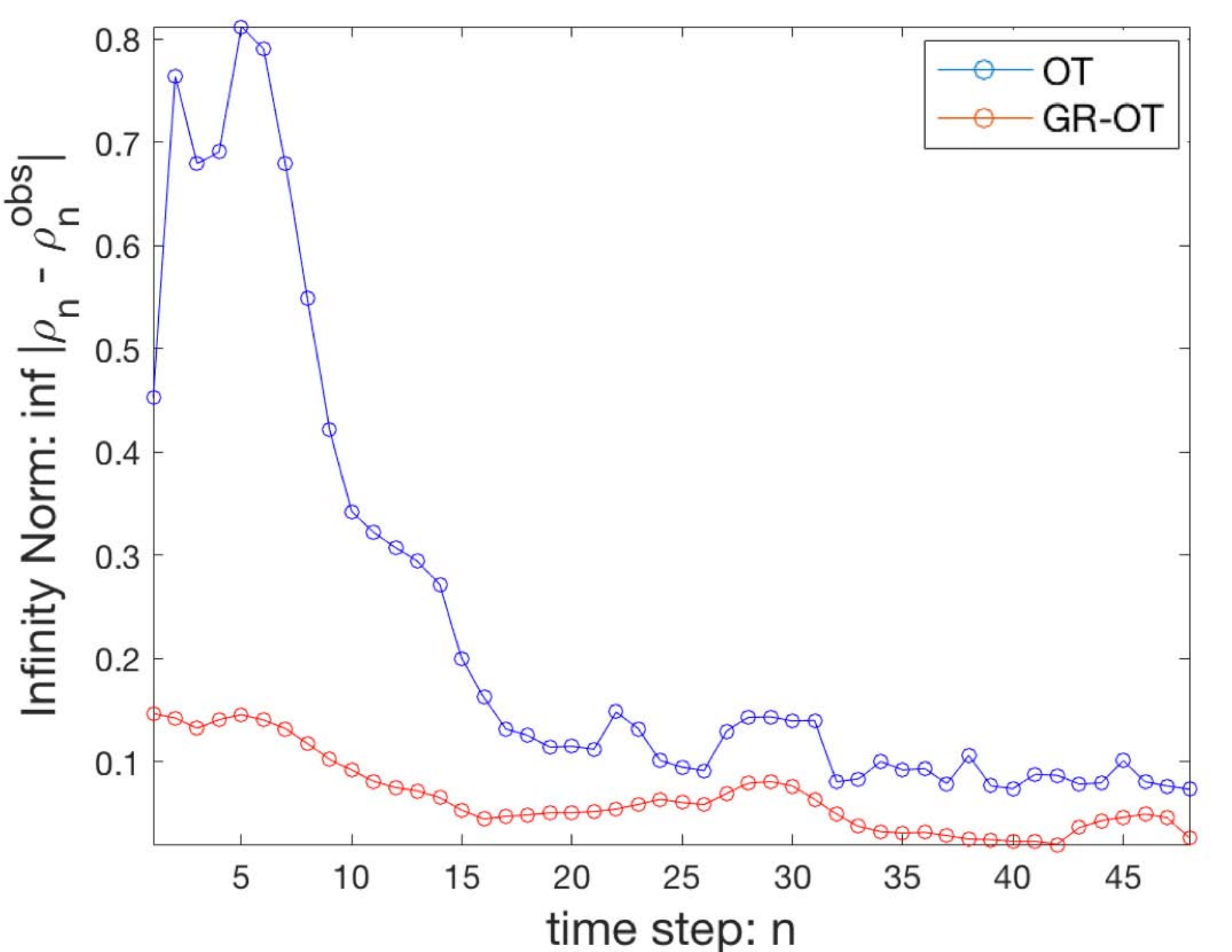}\\
Mean-squared error & Infinity norm of error
\end{tabular}
\end{center}
\caption{Registration error between model returned final density and target image density for the traditional OT model \cite{v1} (shown in blue) and our GR-OT model (shown in red).}
\label{fig:newVSold}
\end{figure}

\begin{figure*}[t!]
\begin{center}
\begin{tabular}{ccc}
\includegraphics[width=0.25\textwidth]{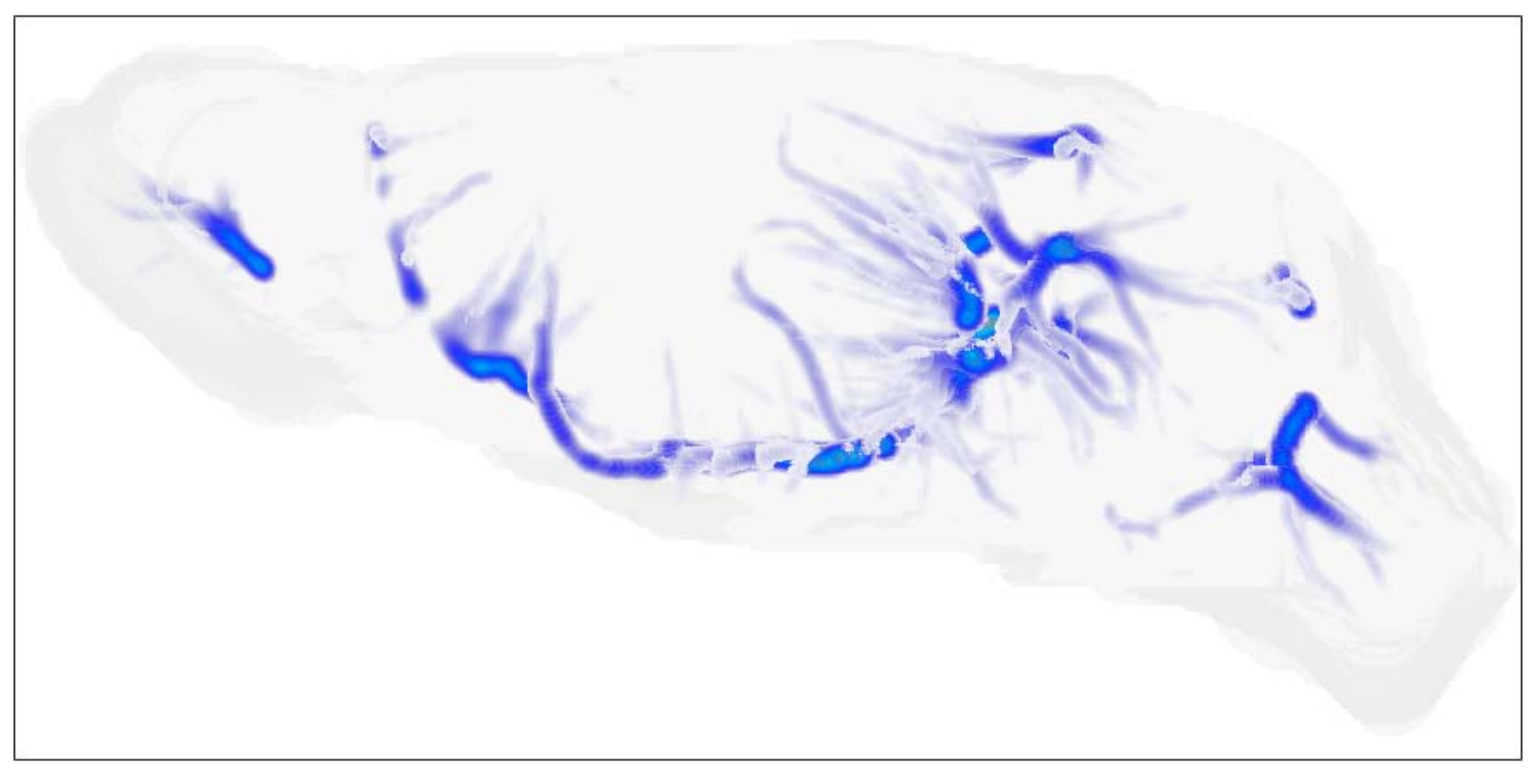}&
\includegraphics[width=0.25\textwidth]{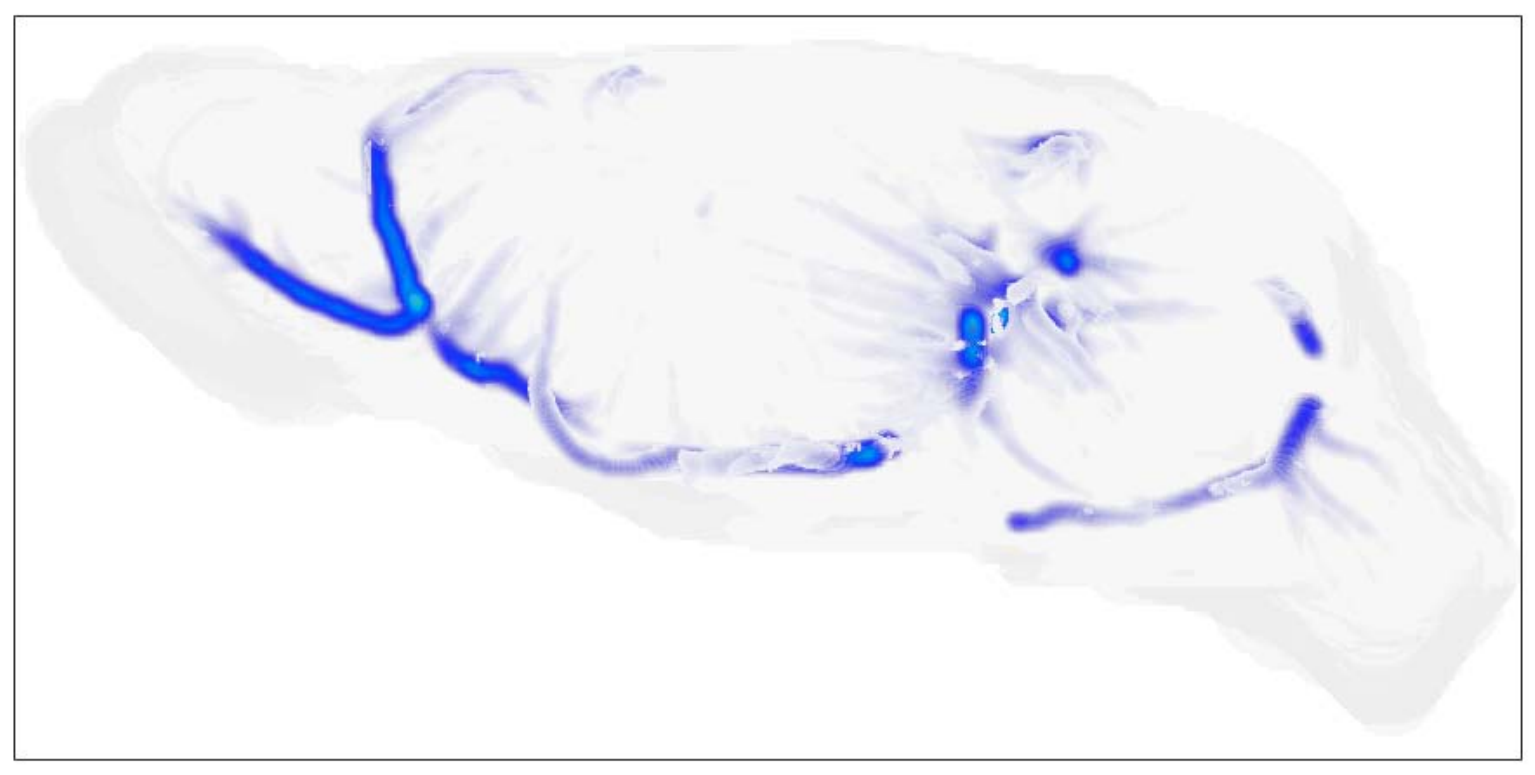}&
\setlength{\tabcolsep}{3pt}
\begin{tabular}[b]{lcccc}
     \hline
     Time step & 1 & 2 & 3 & 4 \\
     \hline
     $\tilde{\sigma}=0.02$: & 0.025  &  0.042  &  0.056  &  0.065  \\
     $\tilde{\sigma}=0.2$: & 0.171  &  0.286  &  0.429 & 0.479 \\
     \hline
     & & & & \\
\end{tabular}\\
(a) & (b) & (c)
\end{tabular}
\end{center}
\caption{Robustness of diffusion parameter. (a) Pathways obtained with $\sigma = 0.002$ and (b) pathways obtained with $\sigma = 0.2$. (c) Root mean square error between `clean' densities obtained with $\sigma = 0.002$ and $\sigma = \tilde{\sigma}$. Top row: $\tilde{\sigma} = 0.02$. Bottom row: $\tilde{\sigma} = 0.2$.
\label{fig:diffusion_param}}
\end{figure*}

In order to quantitatively assess the performance of our model, we look at the registration error between the model returned `clean' density and the target image density. Taking the mean square of the error and the infinity norm of the error, our model (with no diffusion, i.e. $\sigma=0$) yields up to 5 times smaller errors than the traditional OT model proposed in \cite{v1} (Fig.~\ref{fig:newVSold}). This large improvement was possible due to the aforementioned adjustments made to account for noise in the data. We then introduce a little diffusion ($\sigma = 0.002$) and look at the root mean square error between the returned `clean' densities with the `clean' densities obtained by increasing the diffusion parameter by factors of 10 ($\sigma = 0.02, 0.2$). The robustness of the diffusion parameter is shown by these errors, given in Fig.~\ref{fig:diffusion_param}c, as well as by the consistent pathways found with multiple values of $\sigma$, see Figs.~\ref{fig:diffusion_param}a and \ref{fig:diffusion_param}b.


\begin{figure}[t!]
\begin{center}
\begin{tabular}{ccc}
\includegraphics[width=0.32\textwidth]{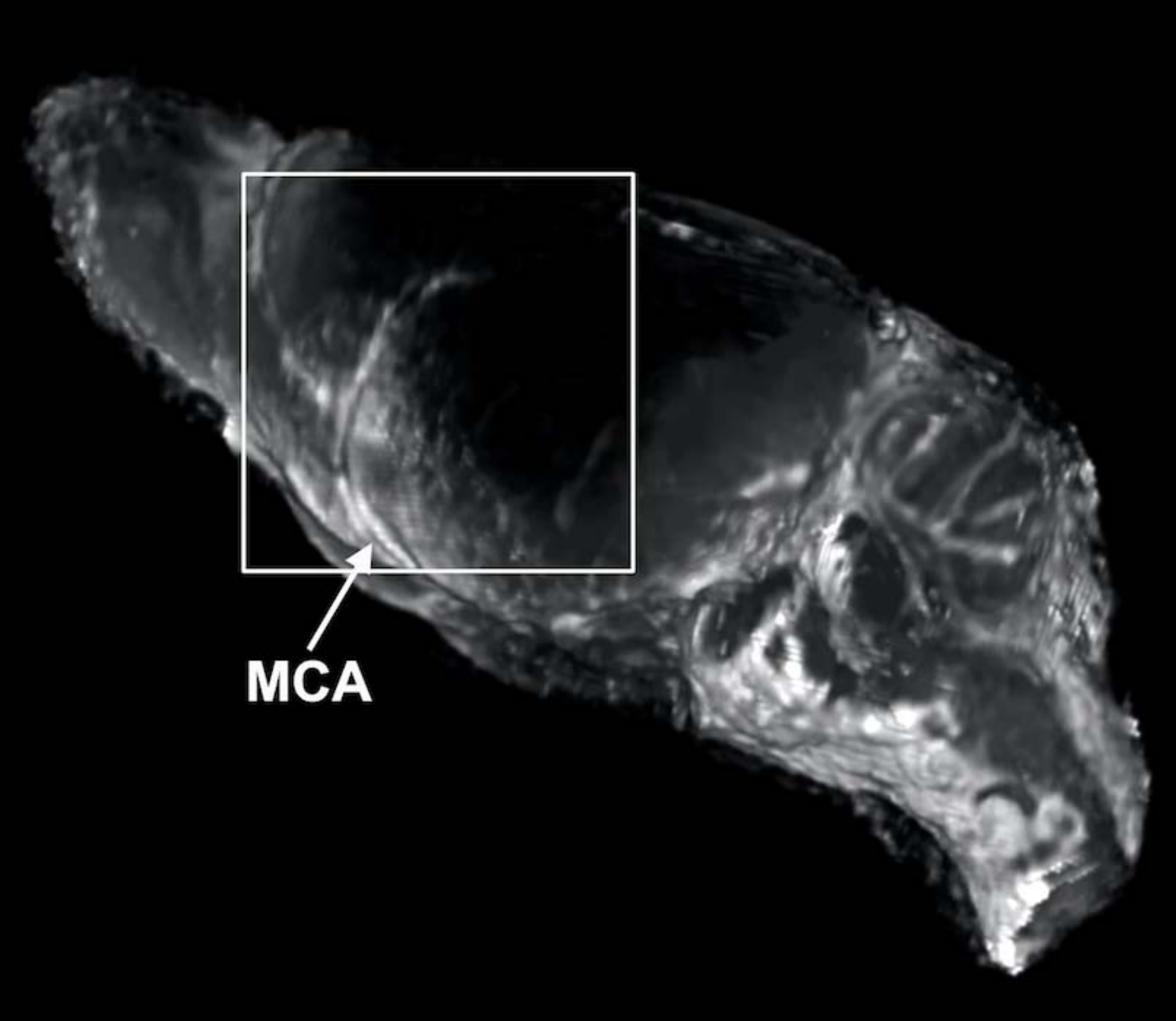}&
\includegraphics[width=0.29\textwidth]{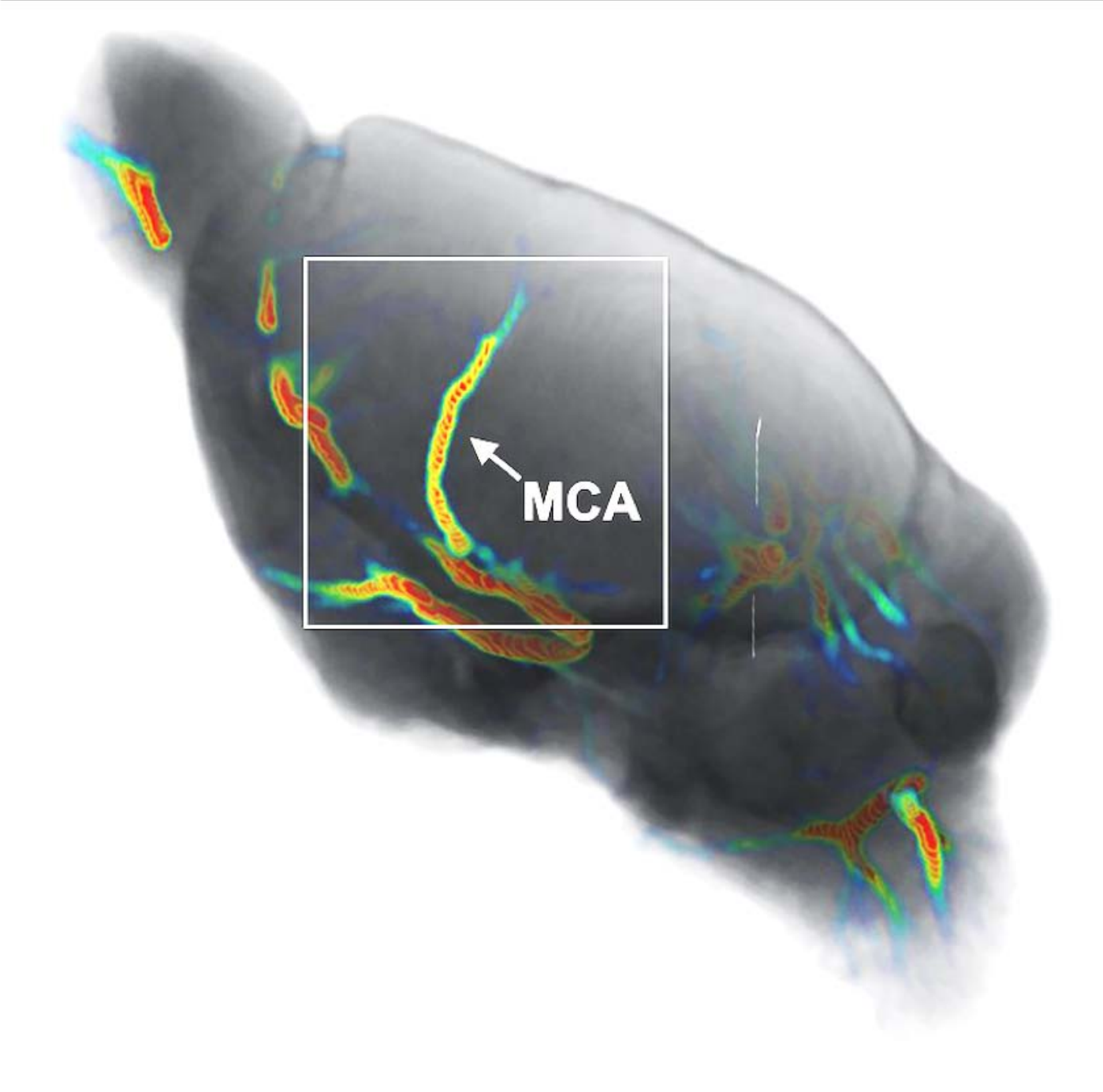}&
\includegraphics[width=0.32\textwidth]{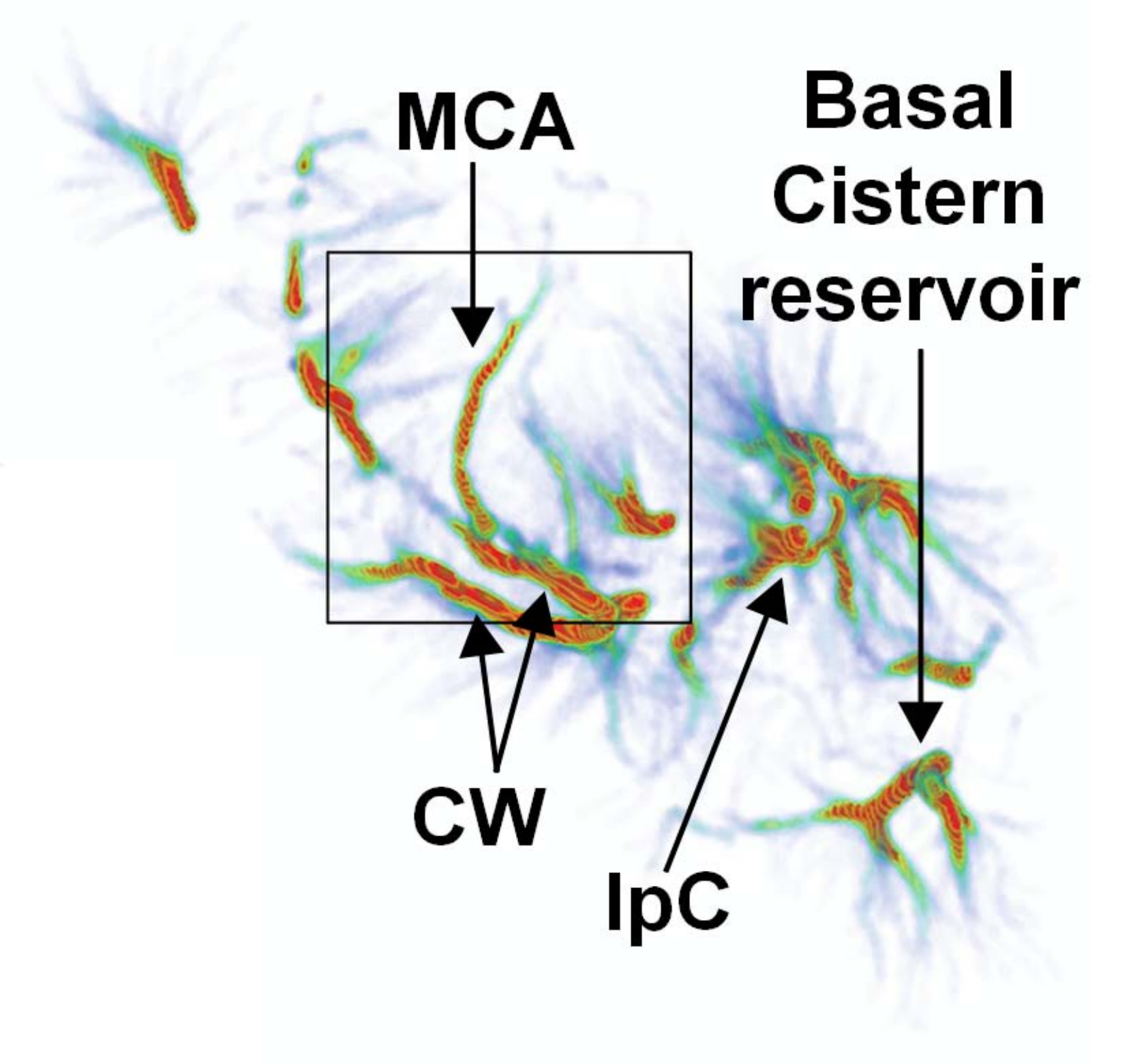}\\
(a) & (b) & (c)
\end{tabular}
\end{center}
\caption{
GlymphVIS pathways.
(a) Original contrast enhanced MRI highlighting the MCA area. (b) 3D volume rendering of the GlymphVIS pathways in relation to the whole rat brain (grey scale, volume rendered) demonstrating that GlymphVIS pathways track CSF transport along the MCA from the level of the Circle of Willis (CW) to where it crosses the olfactory tract (not shown) and proceeds dorsally onto the surface of the brain.  (c) GlymphVIS pathways without the whole brain. Details of the pathways in other areas are now visible including pathway reservoirs associated with the basal cistern, the interpenduncular cistern (IpC) and cleft between the hippocampus and other brain nuclei.}
\label{sld3}
\end{figure}

\begin{figure}[t!]
\begin{center}

\includegraphics[width=\textwidth]{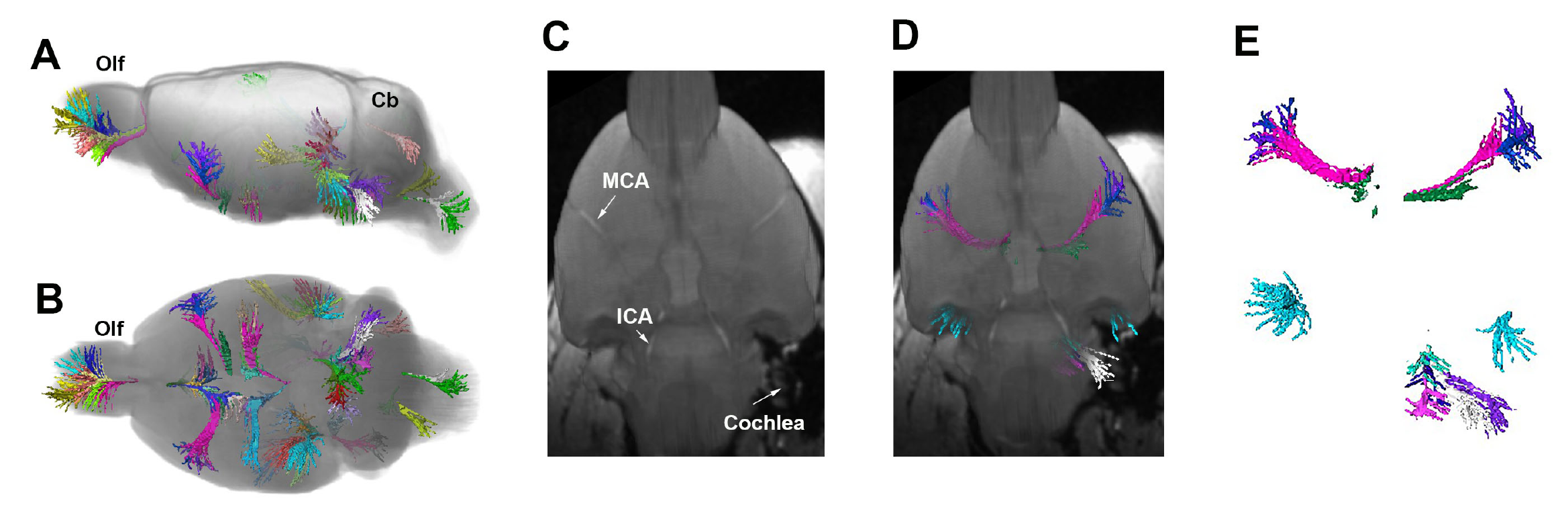}
\end{center}
\caption{GlymphVIS clusters.
(A,B) Clusters shown in different colors (Olf=olfactory bulb and Cb=cerebellum). (C) Anatomical MRI from the ventral surface, where the MCA and internal carotid artery (ICA) can be visualized as single, vascular structures running along the surface of the brain. In addition, the acoustic nerve and inner ear complex (cochlea) is included. (D) Selected streamline clusters related to the MCA and the cochlea overlaid on anatomical template. (E) Selected streamline clusters related to the MCA and the cochlea.}
\label{fig:clusters}
\end{figure}

The utility of GlymphVIS is further validated by its success in reproducing known aspects of glymphatic transport. This is illustrated by the pathways and clusters derived from the MRIs at 1.2hr after contrast infusion into the CSF, shown respectively in Figs.~\ref{sld3} and \ref{fig:clusters}.
In particular, Fig.~\ref{sld3} demonstrates that pathways found by our methodology have accurately captured glymphatic peri-arterial transport along the MCA and in other areas such as the CSF reservoirs. Even more promising, are the trajectories shown by the streamline clusters in Fig.~\ref{fig:clusters}.
This is the first time that specific contrast relevant streamlines have been captured moving towards the inner ear, and illustrates the promise of GlymphVIS and the new GR-OT flow analysis pipeline.


\section{Conclusions and Future Work}

In this paper, we considered a modification of the Benamou-Brenier formulation of OT in which both the continuity and energy cost functionals were modified. This was done to take into account noise as well as possible diffusion in the glymphatic flows for  ``normal'' rat brains. In the future, we also intend to consider cases in which there may be some pathologies, in particular, rat brain models in which there is evidence of AD and vascular dementias. The concept and hypothesis to be tested would be to see if using these mathematical techniques, one could quantitatively differentiate between normal and aberrant CSF flow inside as well as outside the brain, which specifically relate to evolving neuropathology.
Finally, one can consider the technique we have proposed as one of deformable registration. In contrast to other deformable methods such as LDDMM, we are not constrained by only considering diffeomorphic transformations. Moreover, in our setting, we have explicitly taken into account the advection-diffusion nature of the flow, and thus the underlying physics.

%
%

\end{document}